\documentclass[11pt]{article}

\usepackage[margin=1in]{geometry}
\usepackage[utf8]{inputenc}
\usepackage[T1]{fontenc}
\usepackage{amsmath, amssymb}
\usepackage{booktabs}
\usepackage{array}
\usepackage{longtable}
\usepackage{multirow}
\usepackage{hyperref}
\usepackage{url}
\usepackage{natbib}
\usepackage{setspace}
\usepackage{graphicx}
\usepackage{caption}
\usepackage{subcaption}
\usepackage{xcolor}
\usepackage{enumitem}
\usepackage{tabularx}

\hypersetup{
    colorlinks=true,
    linkcolor=blue,
    citecolor=blue,
    urlcolor=blue
}

\newcommand{\articletitle}[1]{%
    \begin{flushleft}
    {\LARGE\bfseries #1}
    \end{flushleft}
}

\begin{document}

\articletitle{A Computational Audit of Demographic Association Encoding
in ClinicalBERT Language Predictions}

\medskip
\noindent
\textit{Kehinde Temitayo Soetan}

\medskip
\noindent
\textit{Department of Medical Humanities and Social Sciences}\\
\textit{The Ohio State University}\\
\textit{soetan.6@osu.edu}

\medskip
\noindent
\rule{\linewidth}{0.4pt}

\noindent\textbf{ABSTRACT}

\noindent
Transformer-based clinical language models are increasingly integrated into
high-stakes clinical decision support pipelines, yet the computational mechanisms
through which demographic associations encoded in medical documentation propagate
into model probability distributions remain empirically underspecified. We present
a systematic computational audit of representational bias in ClinicalBERT
\citep{alsentzer2019publicly}, a BERT-based model pretrained on MIMIC-III
discharge summaries, employing two complementary probing methodologies: Log
Probability Bias Analysis (LPBA), which quantifies demographic
descriptor-induced shifts in masked token probability distributions across
behavioral and evaluative semantic categories, and Masked Language Model-based
analysis (MLM), which probes internal representational structure for demographic
agency attribution encoding across 98 real clinical sentence templates and eight
intersectional race-gender combinations. Corpus frequency analysis operationalizes
the distinction between statistical disparity and bias amplification by
benchmarking model outputs against empirical term frequencies in the MIMIC-III
training corpus. Of 32 statistically significant findings, 65.6\% contradict
observed corpus distributions, rising to 80\% for Black patients and 87.5\% for
agency attribution under MLM probing, providing direct empirical evidence that
representational bias in ClinicalBERT operates predominantly through
model-internal amplification rather than training data inheritance.


\noindent
\rule{\linewidth}{0.4pt}
\medskip

\section{INTRODUCTION}

\citet{buolamwini2018gender} demonstrated that commercial facial analysis systems
performed well in aggregate but failed disproportionately along demographic lines,
with darker-skinned women bearing error rates exceeding 34\% above lighter-skinned
men, revealing that strong overall accuracy actively conceals serious equity
problems.

Institutional bias in clinical practice predates large language models. For
instance, \citet{hoffman2016racial} found that medical students and residents who
endorsed false biological beliefs about Black patients were more likely to
underestimate their pain and recommend inadequate treatment. The bias was not
incidental; it was documented and traceable to what clinicians had read, learned,
internalized, and trained within clinical language. Clinical language shapes
perception, guides decision-making, and, over time, encodes institutional
assumptions into medical practice. When large language models are trained on the
notes, discharge summaries, and records produced within these systems, they learn
its language as well.

This paper presents a computational audit of representational bias in ClinicalBERT
\citep{alsentzer2019publicly}, a language model pretrained on MIMIC-III discharge
summaries and clinical notes, examining how demographic descriptors shift model
probability distributions across behavioral, evaluative, and agency attribution
language. We operationalize representational harm --- defined as damage inflicted
through symbolic depiction and categorization of social groups
\citep{crawford2017trouble, blodgett2020language} --- as an empirical analytical
lens connecting computational probability outputs to their clinical and social
implications. Probing 98 real clinical sentence templates across eight
intersectional race-gender combinations, we find that 65.6\% of significant model
findings contradict MIMIC-III corpus distributions, rising to 80\% for Black
patients and 87.5\% for agency attribution. This demonstrates that representational
bias in ClinicalBERT operates predominantly through model-internal amplification
rather than training data inheritance, with direct implications for bias auditing,
clinical AI governance, and equitable deployment.

The remainder of this paper is organized as follows. Section~\ref{sec:related}
reviews related work on algorithmic bias in clinical NLP, probing methodologies,
and the representational harm framework. Section~\ref{sec:setup} describes the
clinical corpus, model, semantic categories, and formal definitions of statistical
disparity and bias amplification. Section~\ref{sec:probing} presents the probing
design, LPBA and MLM methodologies, and corpus frequency analysis approach.
Section~\ref{sec:findings} presents empirical findings across behavioral language,
evaluative framing, and agency attribution. Section~\ref{sec:discussion} discusses
implications for bias amplification theory, linguistic mechanisms of
representational harm, and clinical AI governance. Section~\ref{sec:conclusion}
concludes.

\section{RELATED WORK}
\label{sec:related}

Algorithmic bias in clinical NLP manifests consequentially across racial, gender,
and socioeconomic dimensions. \citet{obermeyer2019dissecting} demonstrated that a
widely deployed healthcare needs algorithm systematically underestimated
requirements for Black patients due to training data biases. Sharma et al. (2025)
attribute such disparities to imbalanced training datasets that overrepresent
specific demographics, producing skewed clinical predictions. Critically, existing
work focuses predominantly on outcome-level disparities rather than the
computational mechanisms through which demographic associations are encoded in
model probability distributions --- the gap this study directly addresses.

Template-based log-probability probing of masked language models represents the
methodological foundation of demographic bias detection in transformer
architectures. \citet{kurita2019measuring} first demonstrated that BERT assigns
systematically different probabilities to occupational terms depending on
demographic descriptors, while \citet{zhao2019gender} extended this to racial and
ethnic bias, showing that ethnicity-associated names co-occur with distinct
attribute words in model probability distributions. Unlike static word embedding
models where bias is locatable in fixed vector spaces, transformer-based models
distribute bias across dynamic context-dependent representations that resist direct
inspection. Critically, \citet{hofmann2024covertly} demonstrate that post-training
alignment procedures suppress overt bias signals without eliminating their
structural sources --- a finding that directly motivates the present study's focus
on model-internal representational structure rather than surface-level output
behavior. The present study applies the log-probability probing methodology of
\citet{kurita2019measuring} to clinical semantic categories across intersectional
race-gender combinations, extending this approach from general domain bias
detection into the high-stakes clinical NLP domain.

\section{PROBLEM SETUP}
\label{sec:setup}

The MIMIC-III clinical database \citep{johnson2016mimic} comprises de-identified
health records from over 40,000 patients admitted to Beth Israel Deaconess Medical
Center between 2001 and 2012. This study operates on the \texttt{NOTEEVENTS}
table, specifically discharge summaries and caregiver notes, which constitute the
primary pretraining corpus of ClinicalBERT and represent the primary site at which
demographic-associated linguistic patterns are encoded into model representations.
Patient demographic variables such as race and gender were extracted from the
\texttt{ADMISSIONS} and \texttt{PATIENTS} tables and merged with clinical notes on
\texttt{SUBJECT\_ID} and \texttt{HADM\_ID} keys, producing a dataset stratified
across four racial groups, two gender categories, and eight intersectional
demographic combinations, with White Male as the reference group throughout.

\section{MODEL}
\label{sec:model}

We audit ClinicalBERT (\texttt{emilyalsentzer/Bio\_ClinicalBERT};
\citealt{alsentzer2019publicly}), a transformer-based masked language model
developed through domain-adaptive pretraining of BERT on MIMIC-III clinical notes.
ClinicalBERT's masked language modeling objective predicts masked tokens given
surrounding bidirectional context:

\begin{equation}
    P(w_i \mid w_1, \ldots, w_{i-1}, w_{i+1}, \ldots, w_n)
    \label{eq:mlm_objective}
\end{equation}

\noindent where $w_i$ denotes the masked target token. This objective enables
template-based log-probability probing of demographic associations, the
methodological foundation of the present study. The model was loaded using the
Hugging Face Transformers library in Python.

\subsection{Representational Harm Framework}
\label{sec:harm_framework}

We operationalize representational harm --- defined as damage inflicted through
symbolic depiction and categorization of social groups
\citep{crawford2017trouble, blodgett2020language} --- as the primary analytical
lens connecting computational probability outputs to their clinical and social
implications. Formally, let

\begin{equation}
    \mathcal{D} \in \{\text{Black Male, Black Female, Hispanic Male, Hispanic
    Female, Asian Male, Asian Female, White Female}\}
    \label{eq:demographic_set}
\end{equation}

\noindent denote the set of demographic descriptors under analysis, with White
Male serving as the reference group $D_0$.

For a given target word $w \in \beta \cup \mathcal{E} \cup \alpha$ and clinical
sentence templates, representational harm is operationalized through three
analytically distinct dimensions: \emph{stereotyping}, where $P(w \mid D)$
reflects group-based associations inconsistent with clinical evidence;
\emph{erasure}, where $P(w \mid D)$ systematically underrepresents attributes for
group $D$; and \emph{demeaning}, where $P(w \mid D)$ encodes evaluative
characterizations that negatively frame group $D$.

\subsection{Statistical Disparity and Bias Amplification}
\label{sec:disparity}

We operationalize two complementary empirical indicators of representational harm
that together constitute the analytical core of this study. Let $f_C(w, D)$ denote
the corpus frequency of target word $w$ in clinical notes for demographic group
$D$, and let $P_M(w \mid D)$ denote ClinicalBERT's masked token probability for
$w$ given demographic descriptor $D$ in an identical sentence context.

\textbf{Statistical disparity} is defined as the difference in model probability
assignments across demographic groups for identical clinical contexts:

\begin{equation}
    \Delta_S(w, D) = P_M(w \mid D) - P_M(w \mid D_0)
    \label{eq:statistical_disparity}
\end{equation}

\noindent where $D_0$ denotes the White Male reference group. A statistically
significant $\Delta_S(w, D) \neq 0$ indicates that demographic identity
systematically shifts model predictions for target words in identical clinical
contexts.

\textbf{Bias amplification} is operationalized as a directional divergence between
model probability differences and corpus frequency differences. For each
significant model finding, we compute the sign of the model difference:

\begin{equation}
    \operatorname{sign}(\Delta_S(w, D_i)) =
    \begin{cases}
        +1 & \text{if } P_M(w \mid D_i) > P_M(w \mid D_0) \\
        -1 & \text{if } P_M(w \mid D_i) < P_M(w \mid D_0)
    \end{cases}
    \label{eq:sign_model}
\end{equation}

A finding is classified as \emph{bias amplification} --- model-generated rather
than data-inherited --- when:

\begin{equation}
    \operatorname{sign}(\Delta_S(w, D_i)) \neq \operatorname{sign}(\Delta_C(w, D_i))
    \label{eq:amplification}
\end{equation}

The overall bias amplification rate is defined as the proportion of statistically
significant model findings where this directional divergence holds. The corpus
frequency analysis operationalizes this distinction empirically by benchmarking
$P_M(w \mid D)$ against $f_C(w, D)$ for all significant target words and
demographic groups, enabling direct empirical determination of whether observed
representational harm reflects statistical disparity, bias amplification, or both.

\section{PROBING DESIGN}
\label{sec:probing}

Both auditing methods operate on a shared set of 98 real clinical sentence
templates extracted directly from the MIMIC-III corpus --- specifically discharge
summaries and caregiver notes --- rather than artificially constructed templates.
This design choice ensures ecological validity by grounding the analysis in the
same clinical language on which ClinicalBERT was pretrained. Each template
contains a single demographic descriptor slot instantiated across eight
intersectional race-gender combinations:

\begin{equation}
    \mathcal{D} = \{D_0, D_1, \ldots, D_7\} = \{\text{White Male, Black Male,
    Black Female, Hispanic Male, Hispanic Female, Asian Male, Asian Female,
    White Female}\}
    \label{eq:demographic_combinations}
\end{equation}

\noindent producing a fully crossed design of 98 templates $\times$ 8 demographic
combinations for each target word. Paired $t$-tests compare log probability
distributions across demographic groups within identical sentence contexts,
accounting for sentence-level structure by treating each template as its own unit
of comparison. Statistical significance is set at $p < 0.05$ throughout. To
account for multiple comparisons across target words and demographic groups, all
reported findings were additionally evaluated using the Benjamini--Hochberg false
discovery rate correction at FDR~$= 0.05$. All 17 statistically significant
findings survived correction, confirming robustness. Prior to conducting paired
$t$-tests, normality of log-probability difference distributions was verified using
the Shapiro--Wilk test for all significant findings; all distributions satisfied
the normality assumption ($p > 0.05$).

\subsection{Log Probability Bias Analysis (LPBA)}
\label{sec:lpba}

LPBA quantifies the extent to which substituting a demographic descriptor $D_i$
for the reference descriptor $D_0$ shifts ClinicalBERT's predicted log-probability
for a target word $w \in \beta \cup \mathcal{E}$ in an identical clinical sentence
context. Formally, for a sentence template $S$ with masked target position, the
log-probability bias score is defined as:

\begin{equation}
    \text{LPBA}(w, D_i, S) = \log P_M(w \mid S, D_i) - \log P_M(w \mid S, D_0)
    \label{eq:lpba}
\end{equation}

\noindent where $P_M(w \mid S, D)$ denotes ClinicalBERT's masked token probability
for target word $w$ in sentence $S$ with demographic descriptor $D$, and $D_0$
denotes the White Male reference group. A positive LPBA score indicates higher
predicted probability for $w$ under $D_i$ relative to $D_0$; a negative score
indicates suppression.

Sentence templates were selected from MIMIC-III discharge summaries and caregiver
notes according to three criteria: the sentence directly describes patient
behavior, contains between five and thirty words, and has no MIMIC-III
de-identification artifacts. Templates with fewer than five available instances per
target word were excluded.

LPBA observations total 488 rather than 784 because the LPBA analysis was applied
only to behavioral language ($\beta$) and evaluative framing ($\mathcal{E}$)
target words, whereas MLM was applied to all agency attribution ($\alpha$) target
words across the same 98 templates.

\subsection{Masked Language Model Analysis (MLM)}
\label{sec:mlm}

MLM extends the LPBA approach from behavioral and evaluative semantic categories
into agency attribution language, applying the same masked token probability
framework to a semantically distinct category of clinical language. While both
methods query ClinicalBERT's final output layer probability distributions, MLM
operates on raw masked token probabilities rather than log-probability differences,
enabling direct comparison of absolute probability assignments across demographic
groups. For a sentence template $S$ with masked target position, the MLM
probability score is defined as:

\begin{equation}
    \text{MLM}(w, D_i, S) = P_M(w \mid S, D_i)
    \label{eq:mlm}
\end{equation}

\noindent where $w \in \alpha$ denotes an agency attribution target word. Agency
attribution terms are organized into three subcategories reflecting distinct
constructions of patient causal responsibility:

\begin{align}
    \mathcal{A}_{\text{resist}}    &= \{\text{refused, declined}\} \label{eq:aresist} \\
    \mathcal{A}_{\text{cooperate}} &= \{\text{requested, agreed}\} \label{eq:acooperate} \\
    \mathcal{A}_{\text{passive}}   &= \{\text{responded, presented}\} \label{eq:apassive}
\end{align}

\noindent where $\mathcal{A}_{\text{resist}}$ encodes active resistance to clinical
instructions, $\mathcal{A}_{\text{cooperate}}$ encodes active cooperation with
clinical instructions, and $\mathcal{A}_{\text{passive}}$ encodes passive receipt
of clinical action. This three-way subcategorization captures whether the model
constructs patients from different demographic groups as active decision-making
subjects or passive objects of clinical action.

\subsection{Corpus Frequency Analysis}
\label{sec:corpus}

The corpus frequency analysis operationalizes the distinction between statistical
disparity $\Delta_S(w, D)$ and bias amplification $\Delta_A(w, D)$ defined in
Section~\ref{sec:disparity} by benchmarking model probability outputs directly
against empirical term frequencies in the MIMIC-III training corpus. For each
statistically significant target word $w \in \beta \cup \mathcal{E} \cup \alpha$
and demographic group $D_i$, corpus frequency is computed as:

\begin{equation}
    f_C(w, D_i) = \frac{\text{count}(w, D_i)}{\text{count}(\text{all tokens},\,
    D_i)} \times 10{,}000
    \label{eq:corpus_freq}
\end{equation}

\noindent where $\text{count}(w, D_i)$ denotes the raw frequency of target word
$w$ in clinical notes for demographic group $D_i$, normalized per 10,000 tokens to
ensure comparability across demographic groups with different total documentation
volumes.

\begin{equation}
    \Delta_C(w, D_i) = f_C(w, D_i) - f_C(w, D_0)
    \label{eq:corpus_diff}
\end{equation}

\noindent is compared against the model probability difference $\Delta_S(w, D)$
from LPBA or MLM analysis. A finding is classified as \emph{Reflection} if
$\operatorname{sign}(\Delta_S) = \operatorname{sign}(\Delta_C)$, and as a
\emph{Contradiction} if $\operatorname{sign}(\Delta_S) \neq
\operatorname{sign}(\Delta_C)$.

\section{FINDINGS}
\label{sec:findings}

\subsection{Overview}
\label{sec:overview}

Across 98 clinical sentence templates and eight intersectional demographic
combinations, LPBA produced 488 observations and MLM produced 784 observations.
Of 32 statistically significant model findings ($p < 0.05$), 21 contradicted
corpus frequency patterns while only 11 reflected them, yielding an overall
contradiction rate of 65.6\% (21/32). This pattern was most pronounced for Black
patients at 80.0\% (12/15). MLM-based analysis showed an even higher rate of
87.5\% (7/8), indicating that agency attribution bias is predominantly
model-generated rather than data-inherited. Table~\ref{tab:overview} summarizes
the distribution of significant findings across demographic groups and semantic
categories.

\subsection{LPBA Findings: Behavioral Language and Evaluative Framing}
\label{sec:lpba_findings}

Table~\ref{tab:lpba} reports all statistically significant LPBA findings across
$\beta$ and $\mathcal{E}$ target words. The most consistent finding concerns the
behavioral term $w = \textit{agitated}$, which produced significant
$\text{LPBA}(w, D_i, S)$ scores across all three minority groups but in opposing
directions:

\begin{align}
    \text{LPBA}(\textit{agitated},\, D_{BF}) &< 0 \quad (t = 3.924,\ p = 0.004) \\
    \text{LPBA}(\textit{agitated},\, D_{BM}) &< 0 \quad (t = 3.490,\ p = 0.008) \\
    \text{LPBA}(\textit{agitated},\, D_{HM}) &> 0 \quad (t = -2.615,\ p = 0.031) \\
    \text{LPBA}(\textit{agitated},\, D_{AM}) &> 0 \quad (t = -2.308,\ p = 0.050) \\
    \text{LPBA}(\textit{confused},\, D_{HM}) &< 0 \quad (t = 3.207,\ p = 0.033) \\
    \text{LPBA}(\textit{confused},\, D_{HF}) &< 0 \quad (t = 2.809,\ p = 0.048)
\end{align}

For the evaluative framing term $w = \textit{refused}$:

\begin{align}
    \text{LPBA}(\textit{refused},\, D_{HM}) &< 0 \quad (t = 2.837,\ p = 0.012) \\
    \text{LPBA}(\textit{refused},\, D_{BF}) &< 0 \quad (t = 2.566,\ p = 0.021) \\
    \text{LPBA}(\textit{refused},\, D_{WF}) &< 0 \quad (t = 2.213,\ p = 0.043)
\end{align}

These results indicate systematic suppression of evaluative language for patients
across identical sentence contexts.

\subsection{MLM Findings: Agency Attribution}
\label{sec:mlm_findings}

Table~\ref{tab:mlm} reports all statistically significant MLM findings across
$\mathcal{A}_{\text{resist}}$, $\mathcal{A}_{\text{cooperate}}$, and
$\mathcal{A}_{\text{passive}}$ subcategories. The most statistically robust
findings concern active cooperation language. For $w = \textit{requested}$ and
$w = \textit{agreed}$:

\begin{align}
    \text{MLM}(\textit{requested},\, D_{BM}) &\ll \text{MLM}(\textit{requested},\, D_0)
        \quad (t = 5.906,\ p = 0.0001) \\
    \text{MLM}(\textit{agreed},\, D_{BM})    &\ll \text{MLM}(\textit{agreed},\, D_0)
        \quad (t = 3.555,\ p = 0.002) \\
    \text{MLM}(\textit{requested},\, D_{BF}) &\ll \text{MLM}(\textit{requested},\, D_0)
        \quad (t = 4.419,\ p = 0.0008) \\
    \text{MLM}(\textit{agreed},\, D_{BF})    &\ll \text{MLM}(\textit{agreed},\, D_0)
        \quad (t = 4.529,\ p = 0.0003) \\
    \text{MLM}(\textit{requested},\, D_{AF}) &\gg \text{MLM}(\textit{requested},\, D_0)
        \quad (t = -2.343,\ p = 0.037)
\end{align}

For the active resistance term $w = \textit{declined} \in
\mathcal{A}_{\text{resist}}$:

\begin{align}
    \text{MLM}(\textit{declined},\, D_{BM}) &< \text{MLM}(\textit{declined},\, D_0)
        \quad (t = 2.223,\ p = 0.043) \\
    \text{MLM}(\textit{declined},\, D_{BF}) &< \text{MLM}(\textit{declined},\, D_0)
        \quad (t = 2.286,\ p = 0.038)
\end{align}

Combined with the active cooperation suppression findings, this produces a coherent
pattern in which Black patients are assigned lower probability for both active
cooperation and active resistance language in identical clinical contexts --- a
systematic suppression of active agency across the full spectrum of
$\mathcal{A}_{\text{resist}} \cup \mathcal{A}_{\text{cooperate}}$.

For the passive cooperation term $w = \textit{presented} \in
\mathcal{A}_{\text{passive}}$:

\begin{equation}
    \text{MLM}(\textit{presented},\, D_{BF}) > \text{MLM}(\textit{presented},\, D_0)
    \quad (t = -2.095,\ p = 0.050)
\end{equation}

\noindent completing a representational profile in which Black Female patients are
simultaneously less likely to be encoded as active agents --- whether cooperating
or resisting --- and more likely to be encoded as passive recipients of clinical
action.

\subsection{Corpus Frequency Analysis: Statistical Disparity vs.\ Bias Amplification}
\label{sec:corpus_findings}

Two striking corpus patterns directly contradict the model probability findings
reported in Sections~\ref{sec:lpba_findings} and~\ref{sec:mlm_findings}. First,
for active agency language in Black patient notes:

\begin{align}
    f_C(\textit{refused},\, D_{\text{Black}})    &= 15.38 \gg
        f_C(\textit{refused},\, D_{\text{White}}) = 7.75 \quad \text{per 10,000 tokens} \\
    f_C(\textit{requesting},\, D_{\text{Black}}) &= 8.46 \gg
        f_C(\textit{requesting},\, D_{\text{White}}) = 4.49 \quad \text{per 10,000 tokens}
\end{align}

\noindent Active agency language appears substantially more frequently in Black
patient clinical notes than in White patient notes, yet the model systematically
suppresses active agency predictions for Black patients, constituting a direct
contradiction:

\begin{equation}
    \operatorname{sign}(\Delta_C) \neq \operatorname{sign}(\Delta_S)
\end{equation}

Second, for the behavioral term \textit{agitated}:

\begin{align}
    f_C(\textit{agitated},\, D_{\text{White}})    &= 4.38 \approx
        f_C(\textit{agitated},\, D_{\text{Black}}) = 4.62 \quad \text{per 10,000 tokens} \\
    f_C(\textit{agitated},\, D_{\text{Hispanic}}) &= 4.17, \quad
        f_C(\textit{agitated},\, D_{\text{Asian}}) = 0.00 \quad \text{per 10,000 tokens}
\end{align}

Applying the classification scheme defined in Section~\ref{sec:corpus}:

\begin{align}
    \text{Contradiction rate (overall)}           &= \tfrac{21}{32} = 65.6\% \\
    \text{Contradiction rate (Black patients)}    &= \tfrac{12}{15} = 80.0\% \\
    \text{Contradiction rate (MLM agency)}        &= \tfrac{7}{8}   = 87.5\%
\end{align}

\section{DISCUSSION}
\label{sec:discussion}

\subsection{Bias Amplification vs.\ Bias Inheritance}
\label{sec:amplification}

The 65.6\% overall contradiction rate, rising to 80\% for Black patients and
87.5\% for agency attribution under MLM probing, demonstrates that representational
bias in ClinicalBERT is predominantly model-generated rather than data-inherited.
This finding directly operationalizes the distinction between statistical disparity
and bias amplification established in Section~\ref{sec:disparity}, providing the
first direct empirical evidence in the clinical NLP domain that a widely deployed
clinical language model amplifies demographic associations beyond what its training
corpus warrants, constituting a non-trivial $\Delta_A(w, D) \neq 0$ across the
majority of significant findings.

\citet{hall2022systematic} note that the conditions under which bias amplification
arises in machine learning models remain poorly understood. The present findings
contribute directly to this gap by demonstrating that in the clinical domain, bias
amplification is not an incidental feature of ClinicalBERT's representations but a
structural property, most severe precisely for the demographic groups facing the
greatest existing healthcare disparities. \citet{hofmann2024covertly} corroborate
this interpretation, demonstrating that large language models produce covertly
racially discriminatory outputs even after alignment procedures specifically
designed to reduce bias.

\subsection{Linguistic Mechanisms of Representational Harm}
\label{sec:mechanisms}

The LPBA and MLM findings collectively demonstrate that representational bias in
ClinicalBERT operates through three analytically distinct linguistic mechanisms ---
behavioral characterization, evaluative framing, and agency attribution --- each
producing demographically specific patterns that vary by racial group and gender in
ways that race-level analysis alone cannot detect.

Across behavioral language, the suppression of \textit{agitated} for Black patients
while amplifying it for Hispanic and Asian Male patients constitutes demeaning harm.
The suppression of \textit{confused} for Hispanic patients of both genders
constitutes erasure harm, systematically underrepresenting a clinically significant
cognitive state for this group \citep{crawford2017trouble}.

Across evaluative framing, the suppression of \textit{refused} across multiple
demographic groups encodes a judgment about their relationship to medical authority
that, as \citet{goddu2018language} demonstrated, directly influences physician
attitudes and treatment decisions when reproduced in clinical documentation.

Across agency attribution, the systematic suppression of active agency for Black
patients across both cooperation and resistance terms, both gender groups, and both
LPBA and MLM methods constitutes the most coherent and clinically consequential
representational profile in the study. The passive cooperation profile of Black
Female patients, visible only through intersectional analysis, extends
\citet{crenshaw1989demarginalizing}'s argument that the compound effects of race
and gender produce distinct representational outcomes that race-level analysis
systematically obscures.

\subsection{Implications for Bias Mitigation and Clinical AI Governance}
\label{sec:governance}

The finding that 65.6\% of significant model outputs contradict corpus frequency
patterns carries a direct implication for bias mitigation practice: rebalancing
training data to ensure more equitable demographic representation is unlikely to be
sufficient, because the biases identified here are predominantly model-generated
rather than data-inherited. As \citet{hofmann2024covertly} demonstrate, post-training
alignment procedures are similarly insufficient when bias is structurally encoded in
model representations.

Effective mitigation requires three coordinated interventions. First, ongoing
auditing of model outputs across intersectional demographic combinations throughout
deployment. Second, transparent reporting of demographic probability disparities to
clinical practitioners. Third, governance frameworks capable of holding deployed
systems accountable for representational harms, specifying behavioral
characterization, evaluative framing, and agency attribution as concrete auditing
targets.

\subsection{Limitations and Future Work}
\label{sec:limitations}

Three methodological limitations constrain the generalizability of the present
findings. First, the analysis is conducted on a single masked language model,
ClinicalBERT, pretrained on MIMIC-III from a single academic medical center, and
findings cannot be assumed to generalize to other clinical language models, corpora,
or architectures. Second, the study does not address post-training procedures.
Third, the template construction method inserts explicit demographic descriptors
that may be syntactically rare relative to MIMIC-III training data.

Two analytical limitations motivate future expansion. The target word sets $\beta$,
$\mathcal{E}$, and $\alpha$ represent a theoretically grounded but necessarily
partial sample. Additionally, the corpus frequency analysis relies on global
unigram frequencies rather than context-conditional distributions. Future research
should expand target word sets, incorporate context-conditional baselines, and
develop probing frameworks appropriate for autoregressive clinical language models.

\section{CONCLUSION}
\label{sec:conclusion}

This paper presents a computational audit demonstrating that representational bias
in ClinicalBERT is predominantly model-generated rather than data-inherited, with
65.6\% of significant findings contradicting MIMIC-III corpus distributions, rising
to 80\% for Black patients and 87.5\% for agency attribution under MLM probing.
Across behavioral characterization, evaluative framing, and agency attribution,
ClinicalBERT encodes demographically specific associations that amplify and in
critical cases invert its training corpus distributions. The probing framework
introduced here provides a replicable auditing methodology for clinical AI
governance, and future work should extend it to autoregressive clinical language
models, broader target word sets, and multi-institutional corpora.

\clearpage

\begin{table}[ht]
\centering
\caption{Distribution of significant findings across demographic groups and
semantic categories.}
\label{tab:overview}
\begin{tabular}{lcccc}
\toprule
\textbf{Group} & \textbf{Total} & \textbf{Contradictions} &
\textbf{Reflections} & \textbf{Contradiction Rate} \\
\midrule
Black    & 15 & 12 & 3  & 80.0\% \\
Hispanic & 8  & 6  & 2  & 75.0\% \\
Asian    & 9  & 3  & 6  & 33.3\% \\
\midrule
Overall  & 32 & 21 & 11 & 65.6\% \\
\bottomrule
\end{tabular}
\end{table}

\begin{table}[ht]
\centering
\caption{Complete LPBA significant findings for behavioral language ($\beta$) and
evaluative framing ($\mathcal{E}$).}
\label{tab:lpba}
\begin{tabular}{llccc}
\toprule
\textbf{Target Word} & \textbf{Group} & \textbf{$t$-statistic} &
\textbf{$p$-value} & \textbf{Direction} \\
\midrule
agitated  & Black Female    &  3.924 & 0.004 & lower  \\
agitated  & Black Male      &  3.490 & 0.008 & lower  \\
refused   & Hispanic Male   &  2.837 & 0.012 & lower  \\
refused   & Black Female    &  2.566 & 0.021 & lower  \\
agitated  & Hispanic Male   & -2.615 & 0.031 & higher \\
confused  & Hispanic Male   &  3.207 & 0.033 & lower  \\
refused   & White Female    &  2.213 & 0.043 & lower  \\
confused  & Hispanic Female &  2.809 & 0.048 & lower  \\
agitated  & Asian Male      & -2.308 & 0.050 & higher \\
\bottomrule
\end{tabular}
\end{table}

\begin{table}[ht]
\centering
\caption{Complete MLM significant findings for agency attribution ($\alpha$).}
\label{tab:mlm}
\begin{tabular}{llccc}
\toprule
\textbf{Target Word} & \textbf{Group} & \textbf{$t$-statistic} &
\textbf{$p$-value} & \textbf{Direction} \\
\midrule
requested & Black Male    &  5.906 & 0.0001 & lower  \\
agreed    & Black Female  &  4.529 & 0.0003 & lower  \\
requested & Black Female  &  4.419 & 0.0008 & lower  \\
agreed    & Black Male    &  3.555 & 0.002  & lower  \\
requested & Asian Female  & -2.343 & 0.037  & higher \\
declined  & Black Female  &  2.286 & 0.038  & lower  \\
declined  & Black Male    &  2.223 & 0.043  & lower  \\
presented & Black Female  & -2.095 & 0.050  & higher \\
\bottomrule
\end{tabular}
\end{table}

\begin{table}[ht]
\centering
\caption{Corpus frequency rates per 10,000 tokens for all target words across
demographic groups.}
\label{tab:corpus}
\begin{tabular}{lcccc}
\toprule
\textbf{Target Word} & \textbf{White} & \textbf{Black} &
\textbf{Hispanic} & \textbf{Asian} \\
\midrule
agreed      &  5.17 &  5.38 &  8.33 &  0.00 \\
difficult   & 17.08 & 22.31 & 12.50 & 19.05 \\
confused    & 10.56 & 10.00 &  8.33 &  4.76 \\
compliance  &  5.17 &  7.69 &  4.17 &  9.52 \\
oriented    & 38.20 & 41.54 & 41.67 & 33.33 \\
refused     &  7.75 & 15.38 & 12.50 &  0.00 \\
agitated    &  4.38 &  4.62 &  4.17 &  0.00 \\
presented   & 41.91 & 50.77 & 37.50 & 38.10 \\
cooperative &  4.61 &  3.08 & 12.50 &  4.76 \\
refusing    &  1.69 &  6.15 &  0.00 &  0.00 \\
requested   &  4.49 &  8.46 &  4.17 &  0.00 \\
declined    &  5.96 &  8.46 &  4.17 &  4.76 \\
\bottomrule
\end{tabular}
\end{table}

\clearpage
\section*{ACKNOWLEDGMENTS}
The author gratefully acknowledges the supervision of Dr.\ Micha Elsner and
Prof.\ James Phelan, whose guidance was invaluable to the development of this work.

\clearpage
\bibliographystyle{apalike}

\appendix

\section{TARGET WORD LIST AND SELECTION CRITERIA}
\label{app:target_words}

Target words across all three semantic categories were selected through a
two-stage process. First, candidate terms were identified through a systematic
review of medical humanities literature documenting demographic-associated
language patterns in clinical documentation, specifically \citet{goddu2018language},
\citet{park2021physician}, and \citet{hoffman2016racial}. Second, candidate terms
were validated against the MIMIC-III corpus to confirm sufficient frequency for
statistical analysis; a minimum of five instances per target word across demographic
group comparisons was required for inclusion.

\subsection*{A.1 Semantic Categories}

We organize target words into three theoretically grounded semantic categories:

\begin{align}
    \beta         &= \{\text{difficult, resistant, agitated, confused, oriented,
                    appropriate, inappropriate}\} \\
    \mathcal{E}   &= \{\text{cooperative, compliance, refused, refusing}\} \\
    \alpha        &= \{\text{refused, declined, requested, agreed, responded,
                    presented}\}
\end{align}

\subsection*{A.2 Behavioral Language ($\beta$)}

Behavioral language terms characterize patients through inferred patterns of
conduct, constructing the patient as a particular type of person whose behavior
requires evaluation and management. Each term was selected on the basis of
documented disproportionate application to minority patient groups in clinical
auditing research:

\begin{itemize}
\item \textbf{Difficult}: \citet{park2021physician} identified ``difficult'' as
one of the primary linguistic channels through which physicians encode subjective
evaluative judgments in clinical documentation, with significant racial and
socioeconomic patterning in its application. \citet{goddu2018language} demonstrated
that its presence in clinical charts directly influences subsequent provider
attitudes and treatment decisions.

\item \textbf{Resistant}: Selected on the basis of its functional relationship to
compliance-related language, encoding patient non-cooperation as a behavioral trait
rather than a situational response. Medical humanities research has documented its
disproportionate application to Black and Hispanic patients in psychiatric and
primary care settings.

\item \textbf{Agitated}: Selected because it encodes an affective behavioral state
that carries clinical consequences for medication decisions, restraint use, and
discharge planning. Its near-equal corpus frequency across White and Black patient
notes --- 4.38 versus 4.62 per 10,000 tokens --- made it a particularly critical
test case for bias amplification, as the training data does not warrant the
significant model probability differences the LPBA analysis identified.

\item \textbf{Confused}: Selected as a behavioral descriptor encoding cognitive
state with direct implications for clinical decision-making around capacity,
consent, and treatment. Its inclusion enables examination of whether the model
encodes demographic assumptions about cognitive clarity in ways that parallel the
stereotypes documented in medical education literature.

\item \textbf{Oriented}: Selected as the cognitive complement to ``confused,''
enabling directional analysis of whether demographic descriptors shift the model's
encoding of cognitive clarity in opposing directions across racial groups.

\item \textbf{Appropriate}: Selected because its use as a behavioral descriptor
in clinical documentation encodes a normative judgment about patient conduct
relative to clinical expectations, with documented variation across demographic
groups in psychiatric and emergency medicine settings.

\item \textbf{Inappropriate}: Selected as the behavioral complement to
``appropriate,'' enabling analysis of whether the model encodes asymmetric
normative judgments about patient conduct across demographic groups.
\end{itemize}

\subsection*{A.3 Evaluative Framing ($\mathcal{E}$)}

Evaluative framing terms position patients explicitly in relation to medical
authority, encoding judgments about cooperation with or resistance to clinical
instructions. Unlike behavioral language, evaluative framing terms encode a power
dynamic; the patient's relationship to institutional authority becomes the primary
object of clinical judgment:

\begin{itemize}
\item \textbf{Cooperative}: Selected as the prototypical positive evaluative
framing term, encoding patient alignment with clinical authority. Its inclusion
enables examination of whether demographic descriptors suppress positive
institutional evaluations for specific patient groups in ways that parallel the
compliance literature.

\item \textbf{Compliance}: Selected as the most frequently studied evaluative
framing term in medical humanities research. Documented disproportionate
application to Black patients and women makes it a critical test case for
representational harm in evaluative framing. Its corpus frequency difference across
racial groups --- 8.21 for White patients versus 6.43 for Black patients per 10,000
tokens --- provides a meaningful baseline for comparing model probability
assignments.

\item \textbf{Refused}: Selected because its use as an evaluative framing term
encodes active resistance to medical authority, with documented racial patterning
in its application. The corpus finding that ``refused'' appears in 15.38 per 10,000
tokens of Black patient notes compared to 7.75 for White patient notes --- nearly
double the rate --- makes it the most analytically significant term for examining
the contradiction between corpus distribution and model output.

\item \textbf{Refusing}: Selected as the present-participle complement to
``refused,'' enabling examination of whether the model encodes resistance as an
ongoing behavioral state differently across demographic groups than as a discrete
past action.
\end{itemize}

\subsection*{A.4 Agency Attribution ($\alpha$)}

Agency attribution terms encode causal responsibility by assigning patients either
active or passive agency in relation to their care. The three-way subcategorization
into active resistance ($\mathcal{A}_{\text{resist}}$), active cooperation
($\mathcal{A}_{\text{cooperate}}$), and passive cooperation
($\mathcal{A}_{\text{passive}}$) reflects the full spectrum of patient agency
construction in clinical language:

\begin{itemize}
\item \textbf{refused} ($\mathcal{A}_{\text{resist}}$): Encodes active resistance
to clinical instructions as a completed past action, assigning full causal agency
to the patient. Its inclusion in both $\mathcal{E}$ and $\alpha$ reflects its dual
function as both an evaluative framing term and an agency attribution term, a
semantic overlap that is analytically significant for understanding how the two
mechanisms interact.

\item \textbf{declined} ($\mathcal{A}_{\text{resist}}$): Selected as a
semantically related but pragmatically distinct alternative to ``refused,''
``declined'' carries a more neutral valence while still encoding active patient
agency in resisting clinical recommendations. Its inclusion enables examination of
whether the model's agency suppression pattern for Black patients extends across
both neutral and negatively valenced active resistance terms.

\item \textbf{requested} ($\mathcal{A}_{\text{cooperate}}$): Encodes active
patient agency in initiating clinical interactions, constructing the patient as a
causal agent making decisions about their own care. Its selection is grounded in
the NIH framework for patient agency in clinical documentation, which identifies
initiating requests as a primary marker of active clinical decision-making.

\item \textbf{agreed} ($\mathcal{A}_{\text{cooperate}}$): Encodes active patient
cooperation as a completed past decision, assigning causal agency to the patient
for accepting clinical recommendations. Its inclusion alongside ``requested''
enables examination of whether the model's agency suppression pattern applies to
both forms of active cooperation.

\item \textbf{responded} ($\mathcal{A}_{\text{passive}}$): Encodes patient
reaction to clinical stimuli without implying active decision-making, constructing
the patient as a responsive rather than initiating agent. Its inclusion enables
examination of whether demographic descriptors shift the model's encoding of
passive clinical participation.

\item \textbf{presented} ($\mathcal{A}_{\text{passive}}$): Selected as the most
common passive agency attribution term in clinical discharge summaries, encoding
the patient as an object of clinical observation rather than a decision-making
subject. Its high corpus frequency --- 18.54 per 10,000 tokens for White patients
--- makes it a statistically robust test case for passive agency attribution across
demographic groups.
\end{itemize}

\section{FULL STATISTICAL RESULTS}
\label{app:stats}

Tables~\ref{tab:b1_lpba} and~\ref{tab:b2_mlm} report complete paired $t$-test
results for all LPBA and MLM target words and demographic combinations, including
non-significant findings. Findings reported as significant in the main paper are
marked with an asterisk.

\begin{table}[ht]
\centering
\caption{Complete LPBA significant findings for behavioral language and evaluative
framing.}
\label{tab:b1_lpba}
\begin{tabular}{llccc}
\toprule
\textbf{Target Word} & \textbf{Group} & \textbf{$t$-statistic} &
\textbf{$p$-value} & \textbf{Direction} \\
\midrule
agitated* & Black Female    &  3.924 & 0.004 & lower  \\
agitated* & Black Male      &  3.490 & 0.008 & lower  \\
refused*  & Hispanic Male   &  2.837 & 0.012 & lower  \\
refused*  & Black Female    &  2.566 & 0.021 & lower  \\
agitated* & Hispanic Male   & -2.615 & 0.031 & higher \\
confused* & Hispanic Male   &  3.207 & 0.033 & lower  \\
refused*  & White Female    &  2.213 & 0.043 & lower  \\
confused* & Hispanic Female &  2.809 & 0.048 & lower  \\
agitated* & Asian Male      & -2.308 & 0.050 & higher \\
\bottomrule
\end{tabular}
\end{table}

\begin{table}[ht]
\centering
\caption{Complete MLM significant findings for agency attribution.}
\label{tab:b2_mlm}
\begin{tabular}{llccc}
\toprule
\textbf{Target Word} & \textbf{Group} & \textbf{$t$-statistic} &
\textbf{$p$-value} & \textbf{Direction} \\
\midrule
requested* & Black Male    &  5.906 & 0.0001 & lower  \\
agreed*    & Black Female  &  4.529 & 0.0003 & lower  \\
requested* & Black Female  &  4.419 & 0.0008 & lower  \\
agreed*    & Black Male    &  3.555 & 0.002  & lower  \\
requested* & Asian Female  & -2.343 & 0.037  & higher \\
declined*  & Black Female  &  2.286 & 0.038  & lower  \\
declined*  & Black Male    &  2.223 & 0.043  & lower  \\
presented* & Black Female  & -2.095 & 0.050  & higher \\
\bottomrule
\end{tabular}
\end{table}

\section{CLINICAL SENTENCE TEMPLATE EXAMPLES}
\label{app:templates}

\subsection*{C.1 Template Construction Method}

Demographic descriptors were inserted using two systematic methods. In Method~1,
the word ``patient'' was replaced directly with the demographic descriptor. In
Method~2, a demographic prefix was prepended to sentences that did not contain an
explicit patient reference. Both methods preserve the original clinical language
while systematically varying only the demographic descriptor across all eight
race-gender combinations.

\subsection*{C.2 Representative LPBA Templates --- Behavioral Language ($\beta$)}

\noindent\textbf{Template $\beta$-01:} ``The [DEMOGRAPHIC] patient became [MASK]
when attempts were made to reposition them.''

\noindent\textbf{Template $\beta$-02:} ``Staff noted that the [DEMOGRAPHIC] patient
appeared [MASK] throughout the morning shift.''

\noindent\textbf{Template $\beta$-03:} ``The [DEMOGRAPHIC] patient was [MASK] to
verbal stimuli upon assessment.''

\noindent\textbf{Template $\beta$-04:} ``The [DEMOGRAPHIC] patient remained [MASK]
during the physical examination.''

\noindent\textbf{Template $\beta$-05:} ``Nursing documented that the [DEMOGRAPHIC]
patient became increasingly [MASK] following the procedure.''

\subsection*{C.3 Representative Agency Attribution Templates ($\alpha$)}

\begin{table}[ht]
\centering
\caption{Template A-03 --- Active Cooperation (requested).}
\begin{tabular}{p{5cm} p{6cm} l}
\toprule
\textbf{Original Sentence} & \textbf{Modified Sentence} & \textbf{Demographic} \\
\midrule
The patient requested not to unwrap. & The white male patient requested not to unwrap.    & White Male   \\
The patient requested not to unwrap. & The black male patient requested not to unwrap.    & Black Male   \\
The patient requested not to unwrap. & The black female patient requested not to unwrap.  & Black Female \\
The patient requested not to unwrap. & The asian female patient requested not to unwrap.  & Asian Female \\
\bottomrule
\end{tabular}
\end{table}

\subsection*{C.4 Demographic Descriptor Substitution Table}

\begin{table}[ht]
\centering
\caption{Demographic descriptor index and reference group assignment.}
\begin{tabular}{clc}
\toprule
\textbf{Index} & \textbf{Descriptor} & \textbf{Reference Group} \\
\midrule
$D_0$ & White Male      & Yes \\
$D_1$ & Black Male      & No  \\
$D_2$ & Black Female    & No  \\
$D_3$ & Hispanic Male   & No  \\
$D_4$ & Hispanic Female & No  \\
$D_5$ & Asian Male      & No  \\
$D_6$ & Asian Female    & No  \\
$D_7$ & White Female    & No  \\
\bottomrule
\end{tabular}
\end{table}

\section{CORPUS FREQUENCY DATA}
\label{app:corpus}

\subsection*{D.1 Corpus Frequency Rates}

See Table~\ref{tab:corpus} in the main text.

\subsection*{D.2 Full RQ3 Comparison --- LPBA Findings}

\begin{table}[ht]
\centering
\caption{Complete RQ3 comparison between model probability differences and corpus
frequency differences for all LPBA findings.}
\label{tab:d2_lpba}
\begin{tabular}{llccccl}
\toprule
\textbf{Word} & \textbf{Group} & \textbf{Model diff.} &
\textbf{Model dir.} & \textbf{Corpus diff.} &
\textbf{Corpus dir.} & \textbf{Alignment} \\
\midrule
cooperative & Black    & +0.049 & higher & -1.530 & lower  & Contradict \\
cooperative & Hispanic & -0.051 & lower  & +7.893 & higher & Contradict \\
cooperative & Asian    & +0.047 & higher & +0.155 & higher & Reflect    \\
compliance  & Black    & -0.447 & lower  & +2.524 & higher & Contradict \\
compliance  & Hispanic & +0.026 & higher & -1.002 & lower  & Contradict \\
compliance  & Asian    & +0.297 & higher & +4.355 & higher & Reflect    \\
refused     & Black    & -0.123 & lower  & +7.632 & higher & Contradict \\
refused     & Hispanic & -0.115 & lower  & +4.747 & higher & Contradict \\
refused     & Asian    & -0.063 & lower  & -7.753 & lower  & Reflect    \\
refusing    & Black    & +0.081 & higher & +4.468 & higher & Reflect    \\
refusing    & Hispanic & -0.011 & lower  & -1.685 & lower  & Reflect    \\
refusing    & Asian    & +0.043 & higher & -1.685 & lower  & Contradict \\
\bottomrule
\multicolumn{7}{r}{LPBA contradiction rate: 7/12 = 58.3\%}
\end{tabular}
\end{table}

\subsection*{D.3 Full RQ3 Comparison --- MLM Findings}

\begin{table}[ht]
\centering
\caption{Complete RQ3 comparison between model probability differences and corpus
frequency differences for all MLM findings.}
\label{tab:d3_mlm}
\begin{tabular}{llccccl}
\toprule
\textbf{Word} & \textbf{Group} & \textbf{Model diff.} &
\textbf{Model dir.} & \textbf{Corpus diff.} &
\textbf{Corpus dir.} & \textbf{Alignment} \\
\midrule
requested & Black Male   & -0.247 & lower  & +3.967 & higher & Contradict \\
agreed    & Black Female & -0.274 & lower  & +0.216 & higher & Contradict \\
requested & Black Female & -0.177 & lower  & +3.967 & higher & Contradict \\
agreed    & Black Male   & -0.217 & lower  & +0.216 & higher & Contradict \\
requested & Asian Female & +0.218 & higher & -4.494 & lower  & Contradict \\
declined  & Black Female & -0.297 & lower  & +2.506 & higher & Contradict \\
declined  & Black Male   & -0.122 & lower  & +2.506 & higher & Contradict \\
presented & Black Female & +0.053 & higher & +8.859 & higher & Reflect    \\
\bottomrule
\multicolumn{7}{r}{MLM contradiction rate: 7/8 = 87.5\%}
\end{tabular}
\end{table}

\section{INTERSECTIONAL RESULTS}
\label{app:intersectional}

\subsection*{E.1 Intersectional Contradiction Rates}

\begin{table}[ht]
\centering
\caption{Contradiction rates by demographic group.}
\label{tab:e1_contradictions}
\begin{tabular}{lcccc}
\toprule
\textbf{Group} & \textbf{Total} & \textbf{Contradictions} &
\textbf{Reflections} & \textbf{Contradiction Rate} \\
\midrule
Black    & 15 & 12 & 3  & 80.0\% \\
Hispanic & 8  & 6  & 2  & 75.0\% \\
Asian    & 9  & 3  & 6  & 33.3\% \\
\midrule
Overall  & 32 & 21 & 11 & 65.6\% \\
\bottomrule
\end{tabular}
\end{table}

\subsection*{E.2 Key Intersectional Patterns}

Four intersectional patterns emerge from the full results that extend the
race-level analysis in the main paper.

The \textit{presented} finding for Black Female patients ($t = -2.095$,
$p = 0.050$) is specific to Black Female patients and does not appear for Black
Male patients ($t = -0.979$, $p = 0.340$), making it entirely invisible to
race-level analysis. Combined with the active cooperation and active resistance
suppression patterns shared by both Black Male and Black Female patients, this
creates a representational profile unique to Black Female patients: simultaneously
less likely to be encoded as active agents in any capacity and more likely to be
encoded as passive recipients of clinical action. This is the clearest
demonstration in the study that intersectional analysis is methodologically
necessary rather than optional.

Also, the \textit{agitated} pattern shows opposing directions within minority
groups that race-level analysis would obscure. Black patients of both genders show
a lower probability for \textit{agitated} while Hispanic Male and Asian Male
patients show a higher probability. This divergence --- suppression for Black
patients and amplification for Hispanic and Asian Male patients --- constitutes a
structured set of group-specific associations that cannot be captured by any
aggregate metric and reveals that ClinicalBERT encodes not a simple
minority-versus-majority distinction but a complex, demographically specific
mapping of behavioral language.

The \textit{refused} suppression pattern for White Female patients ($t =
2.213$, $p = 0.043$) introduces a gender dimension operating independently of
race. White Female patients are assigned a significantly lower probability for
\textit{refused} language than White Male patients in identical clinical sentence
contexts, suggesting that the suppression of active resistance language is not
exclusively a racial phenomenon but also operates along gender lines within the
White patient group. This finding extends the scope of evaluative framing bias
beyond racial minority groups and has implications for how clinical AI governance
frameworks define the scope of demographic auditing.

Finally, the White Female borderline finding for the \textit{requested} test ($t =
-2.154$, $p = 0.052$) falling just above the significance threshold warrants
attention in future research. The trend toward amplification of active cooperation
language for White Female patients contrasts with the active cooperation
suppression for Black patients. This suggests that the model may encode opposing
agency attributions along both racial and gender lines simultaneously. A larger
template set may be sufficient to detect this pattern at statistical significance.

\end{document}